\documentclass[letterpaper,10pt,conference]{ieeeconf}

\IEEEoverridecommandlockouts
\usepackage{cite}
\usepackage{amsmath,amssymb}
\usepackage{graphicx}
\usepackage[caption=false,font=footnotesize]{subfig}
\usepackage{booktabs}
\usepackage{multirow}
\usepackage{xcolor}
\usepackage{url}
\usepackage{balance}
\usepackage{hyperref}
\usepackage{eso-pic}
\hypersetup{colorlinks=true,linkcolor=black,citecolor=black,urlcolor=blue}
\AddToShipoutPictureFG{%
  \AtPageUpperLeft{%
    \put(\LenToUnit{0.75in},\LenToUnit{-0.40in}){%
      \makebox[\dimexpr\paperwidth-1.5in\relax][l]{%
        \scriptsize\sffamily PREPRINT\hfill
        \thepage}}}}

\title{\LARGE \bf
Scale Anchoring via Implicit Semantics and Sequence-Consistent\\
Loop Closure for VGGT-Based SLAM
}

\author{Ziyi Jin, Kangjin Yu, Beichen Wang, Can Jiang, Zhongpu Xia, and Jie Xu\textsuperscript{*}%
\\
\textit{Anyverse Dynamics}%
\thanks{\textsuperscript{*}Corresponding author.}}

\begin{document}
\maketitle
\thispagestyle{empty}
\pagestyle{empty}

\begin{abstract}
Feed-forward geometric foundation models enable dense monocular reconstruction
without requiring camera calibration, but independently reconstructed submaps
can accumulate scale drift, and even a valid revisit can yield an inconsistent
cross-submap measurement. We extend VGGT-SLAM~2.0 with fixed-root scale
anchoring and short-sequence-assisted direct loop connections. SALAD selects
an image pair between the initial submap and each eligible new submap; joint
VGGT inference then uses learned implicit semantic-geometric cues to supply a
scale-only anchor while preserving the original overlap chain. For loop closure,
neighboring image pairs support a consistency
check and translation-scale alignment. Each accepted candidate contributes one
robust Sim(3) factor between existing center-frame nodes, without an auxiliary
loop submap. A 17-sequence comparison lowers mean Sim(3) ATE on nine TUM
sequences from 0.1018 to 0.0451\,m relative to the VGGT-SLAM~2.0 baseline.
Anchoring improves all seven no-loop module ablations. On three Clio scenes,
a fixed-loop comparison favors the direct connection pipeline over the tested
auxiliary-submap pipeline.
A separate candidate audit exposes additional rejections of visually
overlapping pairs by the short-sequence check. These comparisons distinguish
the gains from long-range scale anchoring and loop connection choices from
the limitations of candidate verification.
\end{abstract}
\noindent\textit{Code:} \href{https://github.com/ziyijin81-afk/scale-anchor-slam}{\scriptsize\texttt{github.com/ziyijin81-afk/scale-anchor-slam}}

\section{Introduction}

Feed-forward visual geometry models simplify dense monocular reconstruction
by predicting cameras and scene geometry directly from images. Their bounded
context, however, requires long streams to be partitioned into independently
reconstructed submaps. VGGT-SLAM~2.0 connects these submaps through a factor
graph~\cite{maggio2026vggtslam2}, extending local reconstruction to longer
trajectories. Two difficulties remain: adjacent-overlap measurements can
accumulate scale drift, and separately reconstructed loop measurements may be
inconsistent with the existing map. The second difficulty is not only a
place-recognition problem: even a true revisit can yield an incompatible
relative scale or local geometry.

\begin{figure}[!t]
    \centering
    \includegraphics[width=0.84\columnwidth]{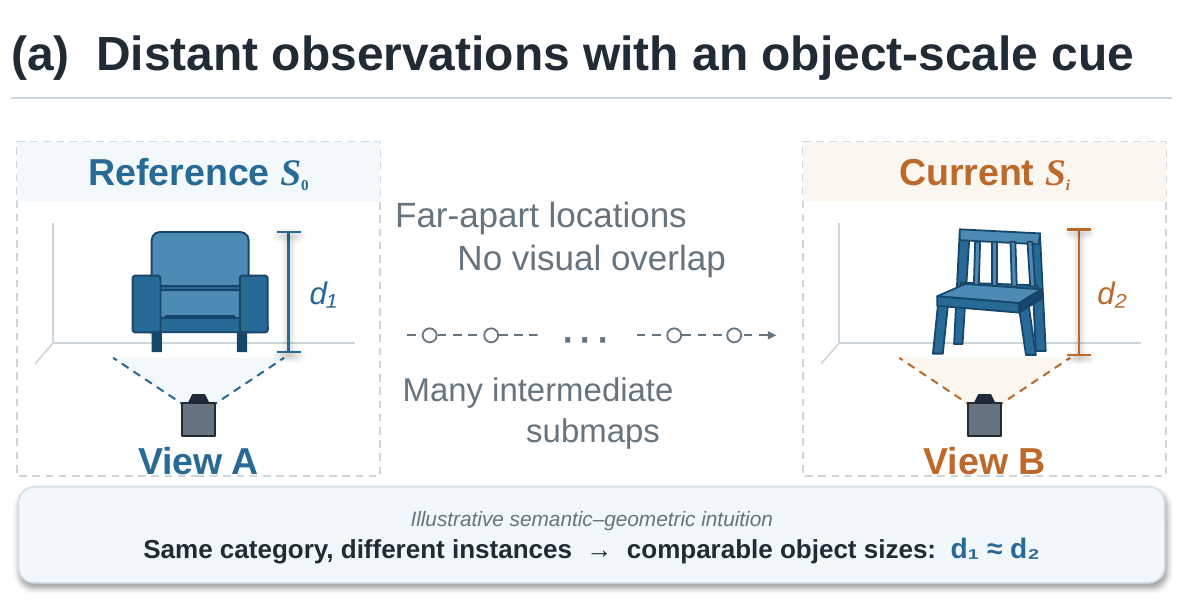}\\[-0.25em]
    \includegraphics[width=0.84\columnwidth]{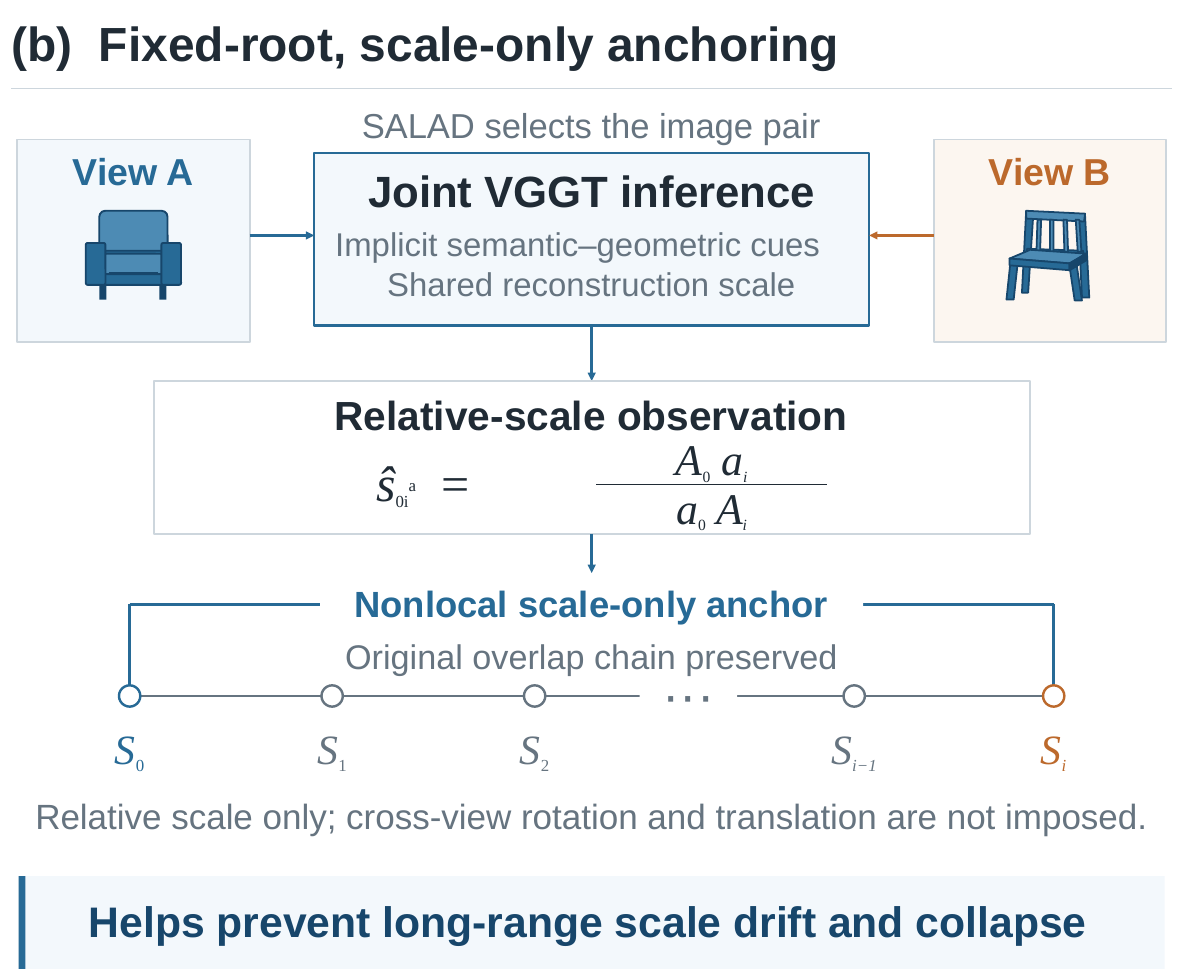}
    \caption{Fixed-root scale anchoring. (a) Learned semantic-geometric cues
    may provide relative-scale information between distant observations;
    objects are schematic. (b) SALAD selects a pair and joint VGGT inference
    supplies a scale-only observation while retaining the overlap chain and
    discarding cross-view pose.}
    \label{fig:scale-anchor-overview}
\end{figure}

Scale drift is particularly relevant when a trajectory traverses new regions
without a verified loop closure. Each overlap provides a local relation, but
the scale of a distant submap depends on a chain of such relations. Small
systematic errors can therefore produce substantial long-range inconsistency
even when neighboring submaps appear well aligned. A loop closure can constrain
this drift only when a suitable revisit is retrieved and accepted. This raises
a complementary question: can learned reconstruction provide useful nonlocal
scale information before a reliable geometric revisit is available?

VGGT offers a potential source of that information. Its DINOv2-initialized
representation and large-scale 3D training support generalization beyond
explicit feature correspondences; the original paper also demonstrates
qualitative reconstruction from non-overlapping views~\cite{wang2025vggt}.
These observations motivate our working hypothesis that implicit
semantic-geometric priors retain useful relative-scale cues even when little
or no covisibility makes relative pose estimation unreliable. In this view,
learned regularities in scene appearance and structure may inform scale
without establishing the precise spatial relationship between the cameras.
An inaccurate cross-view pose therefore need not make all scale information
unusable. The distinction matters because applying an uncertain full-pose
constraint would impose more information than the scale cue alone supports.
Fig.~\ref{fig:scale-anchor-overview} summarizes this motivation and the
resulting scale-only construction.

We consequently exploit scale separately from pose, without assuming that
arbitrary non-overlapping views yield correct scales. Since VGGT uses
scale-normalized training targets~\cite{wang2025vggt}, our target is relative
submap scale, not absolute metric scale. Here, semantic awareness refers to learned scene cues used for image selection and potentially supporting scale inference, without explicit semantic labels.

Our fixed-root anchor uses SALAD to select a compatible image pair between
the current submap and submap zero. Joint VGGT inference places the two images
in a shared reconstruction; comparing camera-frame point norms with their
original submap reconstructions then supplies a relative-scale observation.
The descriptor selects the images rather than measuring scale, and the
estimated cross-view rotation and translation are not inserted as constraints.
Connecting eligible submaps to the same initial reference reduces dependence
on sequential overlap-scale propagation. Unlike selecting an arbitrary recent
reference, the fixed root keeps the reference endpoint fixed instead of moving
it along the growing scale chain. It is a common reference, not a guarantee that the first reconstruction
has the correct metric scale. Weak, robust anchor factors accommodate
potentially biased observations while preserving the original overlap chain.
The anchor does not assert a return to the initial location: it seeks scale
consistency without requiring a verified revisit.

Loop closure addresses a different requirement: when a revisit is accepted,
its relative measurement must be compatible with the reconstructed submaps.
We use neighboring image pairs to check cross-submap consistency and align
translation scales. These supporting views inform verification and estimation,
but each accepted center pair adds only one direct Sim(3) factor between
existing poses, without an auxiliary loop submap or extra neighboring-pair
edges. This separates the evidence used to assess a loop from the connections
inserted into the graph. Scale anchors and loop factors are thus complementary:
the former supplies nonlocal scale observations without asserting a revisit,
whereas the latter corrects pose using accepted revisits.
The two modules are detailed in Sec.~\ref{sec:method}.

The contributions are:
\begin{itemize}
    \item Fixed-root scale anchoring guided by learned retrieval, preserving
    the original local overlap chain.
    \item Short-sequence-assisted direct loop connections that separate
    supporting views from graph edges.
    \item Reference-policy, weight, and module ablations, plus separate
    fixed-candidate and fixed-loop comparisons of verification and connection.
\end{itemize}

We evaluate one configuration on 17 sequences, separating module contributions
on seven sequences and auditing loop acceptance independently of trajectory
accuracy. Per-sequence regressions remain part of the reported evaluation.

\section{Related Work}

\subsection{Feed-Forward Geometry for Dense SLAM}

DUSt3R predicts pairwise pointmaps without known calibration or poses and
aligns them for multi-view reconstruction~\cite{wang2024dust3r}. VGGT jointly
predicts cameras, depth, pointmaps, and tracks from variable-sized image
sets~\cite{wang2025vggt}. Long-stream systems use persistent spatial
memory~\cite{wang2024spann3r}, learned overlapping-clip
registration~\cite{liu2024slam3r}, or real-time SLAM built on two-view
priors~\cite{murai2024mast3rslam}.

VGGT-SLAM uses $\mathrm{SL}(4)$ submap alignment to model projective
ambiguity~\cite{maggio2025vggtslam}. VGGT-SLAM~2.0 reduces graph dimension,
removes planar degeneracies, and verifies loops with VGGT
attention~\cite{maggio2026vggtslam2}. VGGT-SLAM++ combines DEM patches,
DINOv2 features, covisibility, and local optimization~\cite{mandal2026vggtslampp};
VGGT-Long uses overlapping chunks and loop optimization~\cite{deng2025vggtlong}.
Concurrent VGGT-Align derives scale constraints from scene geometric
invariants~\cite{zhang2026vggtalign}. Our root constraints instead use jointly
reconstructed retrieved pairs, without assuming invariant camera height or
road width, while preserving VGGT-SLAM~2.0's local overlap mechanism.

\subsection{Learned Place Recognition and Scale Anchoring}

Sim(3) loop optimization addresses monocular scale drift~\cite{strasdat2010scaledrift};
we seek scale observations even without accepted loops. For image retrieval,
NetVLAD learns VLAD-style aggregation~\cite{arandjelovic2016netvlad}, while
DINOv2 learns transferable visual representations~\cite{oquab2023dinov2}.
SALAD aggregates DINOv2 features with optimal transport and a dustbin for
uninformative features~\cite{izquierdo2024salad}. Descriptor similarity is not
a scale measurement: SALAD selects our frame pair, and joint VGGT
reconstruction supplies its relative scale. Unlike a loop constraint, the
anchor does not assert a shared camera pose.

\subsection{Loop-Closure Verification and Robust Optimization}

To address perceptual aliasing, ORB-SLAM3 combines geometric verification and
covisible observations during map reuse~\cite{campos2021orbslam3}.
SeqSLAM matches coherent sequences under appearance changes~\cite{milford2012seqslam}.
Switchable constraints estimate loop-edge activation~\cite{sunderhauf2012switchable},
whereas dynamic covariance scaling reweights constraints to reduce false-association
effects~\cite{agarwal2013dcs}.

Building on VGGT-SLAM~2.0's attention test, we use neighboring pairs to verify
cross-submap consistency and align translation scales. Each accepted center
pair connects existing poses directly, without an auxiliary loop submap.
The contribution concerns measurement verification, alignment, and connection,
not a new Sim(3) residual or robust loss. We evaluate candidate acceptance
separately from trajectory accuracy and connection quality.

\section{Method}
\label{sec:method}

\subsection{System Overview}

The front end selects a new keyframe when the image displacement reaches
50 pixels. A completed ordinary submap contains 16 new keyframes and one frame
shared with the previous submap; hence a normal VGGT inference contains 17
images. We retain the original VGGT-SLAM~2.0 overlap measurement. For the
shared image, let $\mathbf{X}_{i,p}$ denote a point reconstructed in submap $i$
and transformed into the preceding submap coordinates, and let
$\mathbf{X}_{i-1,p}$ be the corresponding reconstruction in submap $i-1$.
The measured scale is
\begin{equation}
    \hat{s}^{o}_{i-1,i}=
    \underset{p\in\mathcal{V}_{i-1,i}}{\operatorname{median}}
    \frac{\lVert\mathbf{X}_{i-1,p}\rVert_2}
         {\lVert\mathbf{X}_{i,p}\rVert_2},
    \label{eq:overlap-scale}
\end{equation}
where $\mathcal{V}_{i-1,i}$ contains pixels with valid geometric support in the
two reconstructions.

\begin{figure}[!t]
    \centering
    \includegraphics[width=\columnwidth]{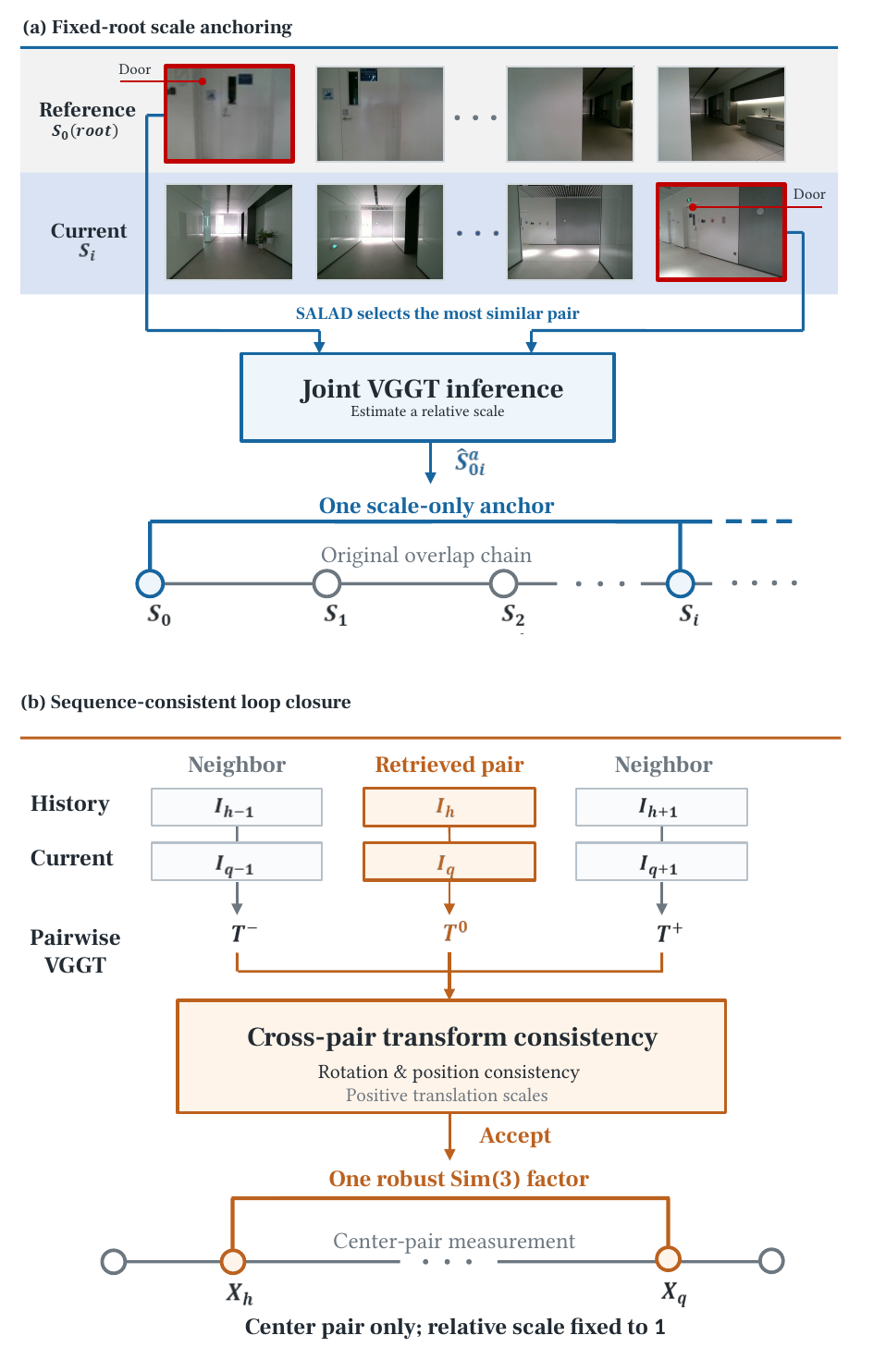}
    \caption{Proposed modules. (a) SALAD selects a pair between fixed root
    $\mathcal{S}_0$ and current submap $\mathcal{S}_i$; joint VGGT inference
    supplies a scale-only anchor alongside the original overlap chain.
    (b) Three pairs check consistency and align translation scales. Each
    accepted center pair adds one robust Sim(3) factor; neighbors do not
    become separate loop edges. Neighbor indices are schematic.}
    \label{fig:method-modules}
\end{figure}

The system retains the original SL(4) pose graph and augments its scale handling
with a scalar GTSAM factor graph. Each ordinary submap has one scale variable
$S_i$, with $S_0=1$ fixing the gauge. Overlap and anchor observations constrain
these scalar variables. Verified loop closures are subsequently handled by a
Sim(3) correction graph over the ordered camera poses.
Fig.~\ref{fig:method-modules} illustrates the fixed-root scale anchor and the
sequence-consistent loop-closure module.

\subsection{SALAD-Guided Fixed-Root Scale Anchor}

Let $\mathcal{S}_0$ (submap zero) be the fixed root submap and $\mathcal{S}_i$ a newly
completed submap. We begin attempting anchors at $i=6$ and subsequently attempt
one anchor for every completed submap. The reference is always $\mathcal{S}_0$:
there is no descriptor threshold and no fallback to another historical submap.
SALAD ranks all image pairs between $\mathcal{S}_0$ and $\mathcal{S}_i$ and
selects the pair with the smallest descriptor distance. Thus the learned
descriptor supplies scene-level, semantically informative selection within a
fixed submap pair, while the scale measurement itself remains geometric.

The selected pair is processed by one two-image VGGT inference. For each image,
we compare camera-frame point norms from its original submap inference and the
joint two-image inference at the same valid pixels. Let $A_0$ and $A_i$ denote
the corresponding robust median norms in the original submaps, and let $a_0$
and $a_i$ denote those in the joint inference. We compute the anchor
measurement as
\begin{equation}
    \hat{s}^{a}_{0i}=\frac{A_0 a_i}{a_0 A_i}.
    \label{eq:anchor-measurement}
\end{equation}
Only finite, positive reconstructions with sufficient valid support are used.
These are necessary computability conditions: we apply no additional learned,
geometric-variance, or statistics-based quality gate to the fixed-root anchor.

The resulting one-dimensional residual is
\begin{equation}
    r^{a}_{0i}=S_i-\hat{s}^{a}_{0i}S_0,
    \label{eq:anchor-residual}
\end{equation}
where $S_i$ denotes the scale state of submap $i$. The anchor constrains only
relative submap scale; it does not impose an absolute pose or a full SL(4)
constraint. The residual is robustified with a Huber kernel. We use ordinary
overlap factor weight $w_o=1$ and anchor factor weight $w_a=0.1$ for every
dataset. The same preselected weight is used across all sequences without sequence-specific tuning.

\subsection{Short-Sequence Loop Verification}

Given a retrieved pair between current submap $i$ and historical submap $j$,
we first require a SALAD descriptor distance below $0.95$ and exclude the three
most recent ordinary submaps from retrieval. This temporal exclusion prevents
local redundant connections from being treated as loop closures. The system
then evaluates three temporally corresponding image pairs: the retrieved
center pair and two neighboring pairs selected within a two-frame radius. Both
forward and reversed temporal orders are considered, and the order with the
lower mean SALAD distance is retained. Each neighboring pair is independently
processed by the same two-image VGGT procedure, while the center pair reuses the
initial loop computation.

For pair $k\in\{1,2,3\}$, let $(R_k^q,t_k^q)$ and $(R_k^h,t_k^h)$ be
submap-to-camera poses in the current and historical submap units available
at verification, with camera centers $c_k^v=-(R_k^v)^Tt_k^v$ for $v\in\{q,h\}$.
The independent two-image reconstruction gives
$D_k=E_k^h(E_k^q)^{-1}=[\widetilde R_k,d_k;0^T,1]$, mapping the query
camera to the historical camera; $E_k^v$ maps that pair's reconstruction
coordinates to camera $v$. Each $d_k$ has its own unknown scale. The common
query-to-history rotation is
\begin{equation}
\begin{aligned}
 R_k &= \Pi\!\left((R_k^h)^T\widetilde R_k R_k^q\right),\\
 \widehat R &= \Pi\!\left(\sum_{k=1}^{3}R_k\right),
\end{aligned}
\label{eq:loop-common-rotation}
\end{equation}
where $\Pi(M)=U\operatorname{diag}(1,1,\det(UV^T))V^T$ for the SVD
$M=U\Sigma V^T$. Define $b_k=(R_k^h)^Td_k$. We estimate the common
translation $t$, scale $s$, and three pairwise translation gauges $\gamma_k$ by
\begin{equation}
 (\widehat t,\widehat s,\widehat\gamma)
 =\underset{t,s,\gamma}{\operatorname{argmin}}
 \sum_{k=1}^{3}\left\|t+s\widehat R c_k^q-\gamma_k b_k-c_k^h\right\|_2^2.
 \label{eq:loop-gauge-fit}
\end{equation}
This is an unconstrained $9\times7$ linear least-squares system, solved by SVD;
we subsequently require rank seven, finite values, $\widehat s>0$, and all
$\widehat\gamma_k>0$. Let $e_k$ be its fitted residual and
$\sigma_v^2=\frac13\sum_k\|c_k^v-\bar c^v\|^2$, where $\bar c^v$ is the mean
center. The normalized position error is
\begin{equation}
 e_p=\frac{\sqrt{\frac13\sum_k\|e_k\|^2}}
 {\max(\sigma_h,|\widehat s|\sigma_q,10^{-12})}.
 \label{eq:loop-position-check}
\end{equation}
We require $e_p\leq0.20$ and rotation RMSE at most $10^\circ$, computed from
the geodesic angles between $\widehat R$ and $R_k$. Every pair must also pass
the original image-match test at $0.95$. Neighbor indices exclude duplicate
center pairs at boundaries. These checks ensure validity, not accuracy.

\subsection{Sequence-Verified Direct Sim(3) Loop Factor}

Only the center pair, indexed by $\star$, becomes a graph edge. The common
rotation in Eq.~\eqref{eq:loop-common-rotation} and center translation from
Eq.~\eqref{eq:loop-gauge-fit} define
\begin{equation}
\begin{aligned}
 \widehat R_{hq} &= \Pi\!\left(R_\star^h\widehat R(R_\star^q)^T\right),\\
 \widehat t_{hq}^{\mathrm{now}}
 &=\frac{S_j^{\mathrm{now}}}{S_j^{\mathrm{est}}}
   \widehat\gamma_\star d_\star .
\end{aligned}
\label{eq:loop-direct-measurement}
\end{equation}
Here $S_j^{\mathrm{est}}$ is the historical submap scale used during verification;
$S_j^{\mathrm{now}}$ is its current optimized scale. We store
$\widehat\gamma_\star d_\star/S_j^{\mathrm{est}}$ in immutable local units and
multiply by $S_j^{\mathrm{now}}$ after every scale update. The loop observation is
$\hat Z_{hq}=\operatorname{Sim}(3)(\widehat R_{hq},\widehat t_{hq}^{\mathrm{now}},1)$:
the fitted $\widehat s$ checks submap consistency, while the factor's measured
relative scale is set to one after scalar-scale optimization. Its seven-dimensional
residual still constrains scale and can change node scales. GTSAM's
\texttt{BetweenFactorSimilarity3} uses standard deviations $0.05$ and Huber
parameter $1.345$. Neighboring pairs support the estimate without becoming
additional loop observations or auxiliary nodes.

\subsection{Staged Graph Optimization}

The scale graph retains the original overlap residual
\begin{equation}
    r^{o}_{i-1,i}=S_i-\hat{s}^{o}_{i-1,i}S_{i-1}
\end{equation}
and the fixed-root anchor residual in Eq.~\eqref{eq:anchor-residual}.
Both use robust Huber losses. The initial scale is fixed to $S_0=1$, and GTSAM
iSAM2 updates the remaining submap scales. The original SL(4) trajectory
construction is then updated with the optimized scales.

When verified loop factors are present, a separate batch Sim(3) correction graph
is built over the scale-corrected camera poses. Adjacent-pose factors preserve the
locally reconstructed trajectory, the first pose fixes the gauge, and each
accepted loop contributes the single direct factor described above. The method
therefore preserves the original local overlap chain, adds long-range scale-only
anchors to the scalar graph, and uses loop closure only for verified global pose
correction. These are staged optimizations rather than one joint objective
over scalar and pose states.

Let $Y_k\in\mathrm{Sim}(3)$ denote a correction-graph camera state and
$\hat{Z}_{uv}$ a relative measurement. Its residual is
\begin{equation}
    \boldsymbol{r}_{uv}=
    \operatorname{Log}\!\left(\hat{Z}_{uv}^{-1}Y_u^{-1}Y_v\right).
    \label{eq:loop-residual}
\end{equation}
All states are initialized from the scale-corrected poses with unit similarity
scale. Consecutive-node measurements are computed from these initial states.
Writing $\mathcal{E}_c$ for these chain edges and $\mathcal{E}_l$ for accepted
loops, the correction objective, together with a strong first-pose prior, is
\begin{equation}
    \min_{\{Y_k\}}\;
    \sum_{(u,v)\in\mathcal{E}_c}\frac{1}{2}
        \lVert\boldsymbol{r}_{uv}\rVert_{\Sigma}^{2}
    +\sum_{(h,q)\in\mathcal{E}_l}
        \rho_H\!\left(\lVert\boldsymbol{r}_{hq}\rVert_{\Sigma}\right),
    \label{eq:correction-objective}
\end{equation}
where $\Sigma=0.05^2 I_7$, $\lVert r\rVert_{\Sigma}^2=r^T\Sigma^{-1}r$,
and $\rho_H$ is the Huber loss with transition $1.345$. The first-state prior
has standard deviation $10^{-6}$ in each dimension. We run
Levenberg--Marquardt for at most 100 iterations and apply the optimized
corrections to the scale-corrected homographies.

\section{Experiments}

\subsection{Experimental Setup}

We evaluate trajectory accuracy and module contributions separately. All
variants use monocular RGB input without supplied camera calibration, depth,
or IMU measurements. Each sequence is keyframe-filtered once at a 50-pixel
disparity threshold. Every variant then uses the same preselected keyframes.
Ordinary submaps contain 16 new keyframes
and one shared overlap frame, except at the start and end of a sequence.
The global overlap estimator and the 25th-percentile confidence threshold
are unchanged. Ordinary VGGT inference is executed separately for each variant,
while the input images and submap partitions remain matched.

The anchor configuration is fixed across sequences: reference submap zero,
SALAD image-pair selection without a descriptor cutoff, one attempt per
submap starting at index six, overlap weight $w_o=1$, and anchor weight
$w_a=0.1$. Our loop closure uses the short-sequence test and direct
factor in Sec.~III, with a minimum ordinary-submap index gap of four.
Each run uses one NVIDIA L20 GPU and the same software environment.

We distinguish five configurations. \emph{Baseline} is VGGT-SLAM~2.0 with
anchors disabled and its original loop-closure pipeline retained.
\emph{B0} disables both proposed modules; \emph{B1} enables only the anchor;
\emph{B2} enables only the proposed loop module; and \emph{Full} (B3) enables
both. All variants use identical keyframes, submap partitions, overlap
constraints, and optimization settings. B0, B1, B2, and Full isolate the two
proposed modules under these matched conditions. Unless stated otherwise,
the reported values are from one run per sequence.

\subsection{Sequences and Evaluation}

The current collection contains 17 sequences: nine Freiburg~1 sequences from
TUM RGB-D~\cite{sturm2012tum}; Cafe~1-1, Cafe~1-2, and Corridor~1-1 from
OpenLORIS~\cite{shi2020openloris}; Apartment, Cubicle, and Office from
Clio~\cite{maggio2024clio}; one HGE iPhone trajectory from
LaMAR~\cite{sarlin2022lamar}; and ADVIO~02~\cite{cortes2018advio}.
The Clio reference poses come from the recorded COLMAP odometry stream,
so its scores measure agreement with a reconstructed reference rather than
independent motion-capture ground truth. LaMAR and Corridor were also used
during method development; this collection is not presented as a held-out
test set.

Our primary metric is position ATE RMSE after one global Sim(3) alignment,
reported in meters. The same evaluator interpolates reference positions
at estimated timestamps within reference coverage. Successful variants of
each sequence have identical exported timestamp sets. Coverage is 100\%
of selected keyframes except Cafe~1-2 (123/124) and Corridor~1-1 (315/316),
with identical coverage for the compared variants. Dataset means are
unweighted means of per-sequence RMSE, not a pooled frame-level RMSE.
We do not average meter-valued errors across unrelated datasets or turn a
single sequence into a scene average. 

\subsection{Experiment 1A: Full-System Comparison}

\begin{table}[t]
    \caption{Experiment 1A: Sim(3) ATE RMSE [m]. Lower is better; bold marks
    the lower value in each pair. Baseline denotes VGGT-SLAM~2.0 under the common
    16-keyframe evaluation protocol.}
    \label{tab:experiment1-main}
    \centering
    \footnotesize
    \setlength{\tabcolsep}{6pt}
    \begin{tabular}{@{}llrr@{}}
        \toprule
        Dataset & Sequence & Baseline & Full \\
        \midrule
        TUM & 360   & \textbf{0.068} & 0.093 \\
            & desk  & 0.183 & \textbf{0.034} \\
            & desk2 & 0.048 & \textbf{0.046} \\
            & floor & 0.243 & \textbf{0.055} \\
            & plant & 0.038 & \textbf{0.026} \\
            & room  & 0.209 & \textbf{0.052} \\
            & rpy   & 0.031 & \textbf{0.028} \\
            & teddy & 0.072 & \textbf{0.055} \\
            & xyz   & 0.024 & \textbf{0.018} \\
            & Mean (9) & 0.102 & \textbf{0.045} \\
        \midrule
        OpenLORIS & Cafe 1-1 & \textbf{0.149} & 0.254 \\
                  & Cafe 1-2 & 0.760 & \textbf{0.517} \\
                  & Cafe mean (2) & 0.454 & \textbf{0.386} \\
                  & Corridor 1-1 & 12.848 & \textbf{6.468} \\
        \midrule
        Clio$^{\dagger}$ & Apartment & 0.610 & \textbf{0.118} \\
                        & Cubicle & 0.212 & \textbf{0.033} \\
                        & Office & \textbf{1.470} & 1.508 \\
                        & Mean (3) & 0.764 & \textbf{0.553} \\
        \midrule
        LaMAR & HGE iPhone & 18.646 & \textbf{1.404} \\
        ADVIO & 02 & 9.882 & \textbf{6.791} \\
        \bottomrule
    \end{tabular}
    \par\smallskip
    \begin{minipage}{\columnwidth}
        \footnotesize
        $^{\dagger}$Clio uses its recorded COLMAP reference trajectory.
        Means use only the indicated sequence group.
    \end{minipage}
\end{table}

Table~\ref{tab:experiment1-main} compares Full with Baseline under the
same 16-keyframe setting. On TUM, mean ATE decreases from 0.1018 to 0.0451\,m,
a 55.6\% reduction, with improvements on eight of nine sequences. This
comparison uses the common 16-keyframe protocol defined above; the published
VGGT-SLAM~2.0 TUM results use 32-frame submaps~\cite{maggio2026vggtslam2}.
The mean decreases from 0.4545 to 0.3857\,m on the two Cafe sequences, and
from 0.7639 to 0.5530\,m on the three Clio scenes. Corridor~1-1 and ADVIO~02
improve by 49.7\% and 31.3\%, respectively.

Neither Baseline nor Full accepts loops on Corridor 1-1 or ADVIO 02.
Baseline rejects all 10 and 29 retrieved candidates, respectively, at its image-match test.
These failures are consistent with limited cross-view overlap:
Corridor 1-1 contains repetitive corridor structures and substantial viewpoint changes,
while ADVIO 02 provides few strong same-view visual revisits.
Since neither sequence accepts a loop, the observed improvement primarily reflects the effect of scale anchoring.

The gains are not uniform: TUM~360, Cafe~1-1, and Clio Office have higher
ATE with Full. On Office, loop integration offers no improvement over
Anchor-only (1.508 versus 1.500\,m), and diagnostic admission of an excluded
revisit does not recover that accuracy, suggesting limited benefit from the
available cross-submap measurements; Full is also slightly worse than
Baseline (1.470\,m).
Cafe~1-1 produces one anchor and no accepted loop, whereas
TUM desk, rpy, and xyz improve without producing any anchor. Improvements
on those TUM sequences therefore cannot be attributed to the anchor.
For LaMAR, Full reduces ATE from 18.6461 to
1.4041\,m (92.5\%) under the common evaluator.

\subsection{Experiment 1B: Module Contributions}

\begin{table}[t]
    \caption{Experiment 1B: module ablation, Sim(3) ATE RMSE [m].
    B0: neither module; B1: anchor only; B2: improved loop only;
    Full: both. Bold marks the lowest value, including ties; underline marks
    the second-lowest distinct value.
    $L$ is the number of accepted Full loops, not labeled true loops.}
    \label{tab:experiment1-modules}
    \centering
    \footnotesize
    \setlength{\tabcolsep}{3pt}
    \begin{tabular}{@{}lrrrrr@{}}
        \toprule
        Sequence & B0 & B1 & B2 & Full & $L$ \\
        \midrule
        TUM room & 0.088 & 0.085 & \underline{0.056} & \textbf{0.052} & 3 \\
        Corridor 1-1 & \underline{12.848} & \textbf{6.468} & \underline{12.848} & \textbf{6.468} & 0 \\
        LaMAR HGE & 7.948 & \underline{1.622} & 7.778 & \textbf{1.404} & 1 \\
        Clio Apartment & 0.181 & \underline{0.140} & 0.164 & \textbf{0.118} & 1 \\
        Clio Cubicle & 0.073 & 0.043 & \underline{0.042} & \textbf{0.033} & 2 \\
        Clio Office & 2.120 & \textbf{1.500} & 2.156 & \underline{1.508} & 1 \\
        ADVIO 02 & \underline{9.882} & \textbf{6.791} & \underline{9.882} & \textbf{6.791} & 0 \\
        \bottomrule
    \end{tabular}
\end{table}

Table~\ref{tab:experiment1-modules} separates module effects on seven sequences;
Full values are reused from Table~\ref{tab:experiment1-main}, not independent
trials. B0--B1 isolates anchoring, B0--B2 the loop pipeline, and B1--Full the
additional loop benefit after anchoring.

Anchoring lowers ATE on all seven sequences, although TUM room improves only
from 0.0876 to 0.0850\,m. The zero-loop Corridor and ADVIO runs have identical
Baseline/B0/B2 errors and identical B1/Full errors. Full outperforms either
module alone on TUM room, LaMAR, Clio Apartment, and Clio Cubicle.
LaMAR illustrates their complementarity: ATE falls from 7.9485 to 1.6221\,m
with anchoring, then to 1.4041\,m with both modules.

\subsection{Experiment 2: Anchor Design Ablation}
\label{sec:weight-ablation}

\begin{table}[t]
    \caption{Experiment 2: anchor-reference ablation with loop closure
    disabled. Values are Sim(3) ATE RMSE [m]; bold and underline mark the
    best and second-best values per sequence. Fixed-root denotes connection to submap zero.}
    \label{tab:anchor-reference}
    \centering
    \footnotesize
    \setlength{\tabcolsep}{3.2pt}
    \begin{tabular}{@{}lrrrr@{}}
        \toprule
        Reference strategy & Room & Corridor & ADVIO & LaMAR \\
        \midrule
        No anchor & 0.088 & 12.848 & 9.882 & 7.948 \\
        Fixed-root + middle frame & 0.090 & 7.239 & \underline{6.942} & \underline{2.902} \\
        Any history + SALAD & \textbf{0.080} & \underline{7.194} & 9.793 & 8.168 \\
        Fixed-root + SALAD (ours) & \underline{0.085} & \textbf{6.468} &
        \textbf{6.791} & \textbf{1.622} \\
        \bottomrule
    \end{tabular}
\end{table}

\begin{figure}[!t]
    \centering
    \includegraphics[width=\columnwidth]{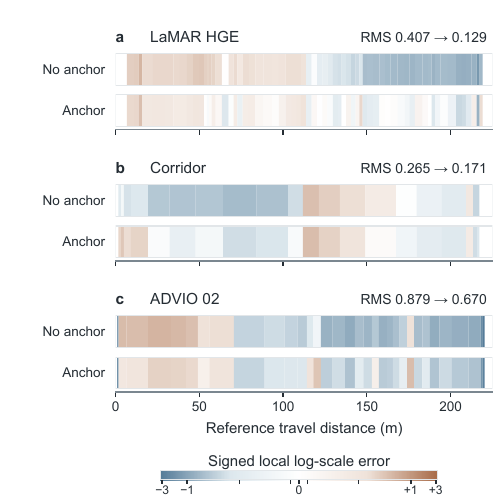}
    \caption{Local scale consistency with loop closure disabled and
    $w_a=0.1$. Each complete trajectory receives one global Sim(3) alignment.
    Each strip encodes the signed local log-scale error at each submap's final
    reference travel distance; blue and muted orange denote negative and
    positive errors, respectively, while white denotes values near zero. The
    upper and lower strips show no anchor and anchor. A shared symmetric-log
    color scale is used across panels. All 64 LaMAR, 20 Corridor, and 34 ADVIO
    submaps are retained, including the initial and final submaps. RMS values
    follow the no-anchor-to-anchor order. These are final, single-run
    trajectories, not causal online estimates.}
    \label{fig:anchor-scale}
\end{figure}

\begin{figure}[!t]
    \centering
    \includegraphics[width=0.70\columnwidth]{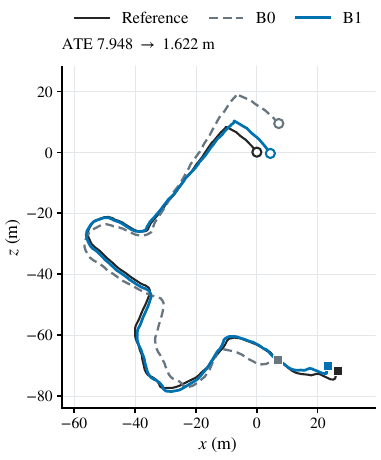}
    \caption{LaMAR HGE trajectories with loop closure disabled after
    per-method global Sim(3) alignment. B0 has no anchor; B1 uses the
    fixed-root anchor. Circles and squares mark the start and end. ATE decreases
    from 7.948 to 1.622\,m.}
    \label{fig:lamar-trajectory}
\end{figure}

We isolate the anchor reference policy on four development sequences with
loop closure disabled. All variants reuse the same keyframes and submap
partition. Anchor-enabled variants have the same opportunities, starting at
submap six and attempting one anchor per subsequent submap, and use
$w_a=0.1$ and $w_o=1$. \emph{Fixed-root + middle frame} selects the temporal middle
frame at both endpoints, using index $\lfloor(N-1)/2\rfloor$ in each submap's
retrieval-frame list of length $N$, and always connects to submap zero.
\emph{Any history +
SALAD} searches all historical submaps separated by at least two submaps,
whereas \emph{Fixed-root + SALAD} restricts the search to image pairs between the
current and root submaps. Neither SALAD strategy uses a descriptor threshold,
distance decay, or span-dependent weighting.

Fixed-root + SALAD improves over No anchor and Fixed-root + middle frame
on all four sequences (Table~\ref{tab:anchor-reference}). It is best on Corridor,
ADVIO, and LaMAR; Any history + SALAD is best on TUM room.
Historical search nearly returns to the no-anchor error on ADVIO and slightly
exceeds it on LaMAR. Thus fixed-root retrieval helps without being universally
optimal.

On Indoor, Rosbag, Corridor, and LaMAR, we additionally evaluated anchor
weights $0$, $0.05$, $0.1$, $0.25$, $0.5$, $1$, and $2$ with
the Fixed-root + SALAD policy and loop closure disabled. The response is
sequence-dependent: $0.25$ further improves Corridor, but degrades Indoor and
LaMAR relative to $0.1$, while changing Rosbag only marginally. We therefore use the preselected value $w_a=0.1$ for all reported experiments without sequence-specific tuning.

\noindent\textit{Local scale consistency.}
Fig.~\ref{fig:anchor-scale} provides a post-hoc scale diagnostic on three
long sequences using the same ordinary-submap predictions and matched scale
graphs. For submap $i$, let $c_i$ be the median ratio of estimated
to reference camera-center distances over unique pose pairs with reference
separation at least $0.1$\,m. We evaluate $e_i=\ln(\alpha c_i)$, where
$\alpha$ is the single global Sim(3) alignment scale of the complete
trajectory; local submaps are not independently rescaled.
RMS decreases on all three sequences, while the residuals retained at the
sequence endpoints show that improvement is not uniform across submaps. Camera-center motion
provides a scale proxy, so local deformation and reference error can also
affect this diagnostic. Fig.~\ref{fig:lamar-trajectory} shows the corresponding
LaMAR trajectory improvement.

\subsection{Experiment 3: Loop Verification and Connection}
\label{sec:loop-ablation}

\begin{table}[t]
    \caption{Loop verification and connection ablation on Clio.}
    \label{tab:loop-ablation}
    \centering
    \footnotesize
    \setlength{\tabcolsep}{4pt}
    \begin{tabular}{@{}lrrr@{}}
        \toprule
        & Apartment & Cubicle & Office \\
        \midrule
        Candidate pool & 2 & 4 & 9 \\
        Single-pair accepted & 2 & 2 & 2 \\
        Short-sequence accepted & 1 & 2 & 1 \\
        \midrule
        Anchor-only & \underline{0.140} & \underline{0.043} & \textbf{1.500} \\
        Anchor + auxiliary-submap loop & 0.505 & 0.175 & 3.370 \\
        Anchor + direct loop (Full) & \textbf{0.118} & \textbf{0.033} & \underline{1.508} \\
        \bottomrule
    \end{tabular}
    \par\smallskip
    \begin{minipage}{\columnwidth}
        \footnotesize
        Top: counts on identical candidates. Bottom: $\mathrm{Sim}(3)$ ATE
        RMSE [$\mathrm{m}$]; bold and underline mark the best and second-best
        ATE. The two loop connections
        use identical accepted loops and $w_a=0.1$, but differ in topology,
        backend, and robust treatment.
    \end{minipage}
\end{table}

\begin{figure}[!t]
    \centering
    \includegraphics[width=0.70\columnwidth]{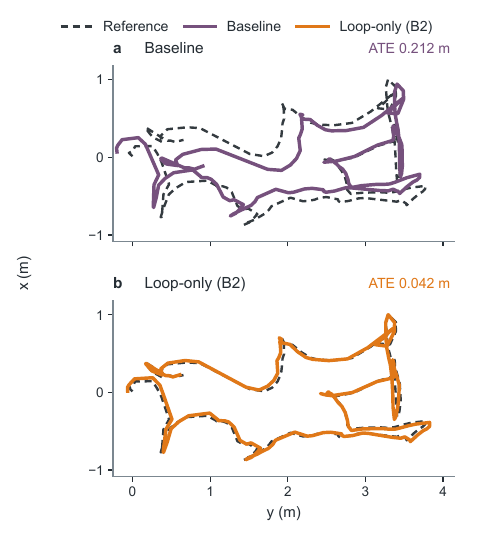}
    \caption{Clio Cubicle trajectories with Anchor disabled: (a) Baseline;
    (b) Loop-only (B2). Both use 152 keyframes and a global Sim(3) alignment
    to the recorded COLMAP reference; axes match and ATE is 3D RMSE.}
    \label{fig:clio-loop-comparison}
\end{figure}

\begin{figure}[!t]
    \centering
    \includegraphics[width=0.86\columnwidth]{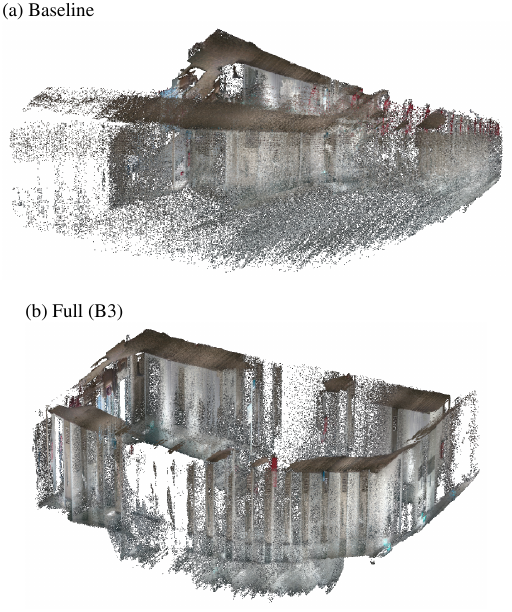}
    \caption{Dense reconstruction on the NTU Indoor sequence from the
    INS dataset~\cite{deng2025mneslam}. (a) Baseline;
    (b) Full (B3). Full reduces the compressed and duplicated structures visible
    in the Baseline map and yields a more coherent corridor layout.}
    \label{fig:indoor-dense-map}
\end{figure}

\noindent\textit{Fixed-candidate audit.}
We evaluate both checks on the same 15 candidates retrieved by the Full
pipeline on Clio: two in Apartment, four in Cubicle, and nine in Office.
Both checks therefore use the same SALAD candidates and minimum submap gap
of four. The inherited single-pair image-match test rejects nine candidates;
the added short-sequence check retains four of the remaining six. These counts apply only to the fixed candidate set. They differ from the end-to-end Baseline counts
in Experiment~1, whose retrieval pool is not fixed to that of Full.

The additional checks assess measurement consistency beyond visual overlap:
one Apartment candidate fails a neighboring image-match test, while one Office
candidate has a normalized position RMSE of $0.468$ (threshold $0.20$) and
negative fitted translation gauges. Both are labeled overlapping revisits by
our provisional visual/reference audit, showing that the checks can also
reject plausible revisits. Because these labels are not independently adjudicated,
we report acceptance counts rather than precision or recall.

\noindent\textit{Fixed-loop connection comparison.}
We next fix the four accepted center-frame pairs: one in Apartment, two in
Cubicle, and one in Office. Each scene is run with the legacy auxiliary-submap
connection and with the direct connection, retaining the same keyframes,
ordinary-submap partition, anchor configuration, and candidate endpoints.
We verify identical exported timestamps, anchor image pairs, and accepted loop
counts. Ordinary VGGT inference is executed separately for each variant, as in
Experiment~1.

The auxiliary-submap connection routes each loop through a two-image submap
using the original SL(4) connection machinery. The direct variant uses the
robust Sim(3) correction graph in Sec.~III. Because the two variants also differ in state representation and robust optimization, this experiment compares the complete connection pipelines rather than the graph topology alone.

Table~\ref{tab:loop-ablation} reuses the direct results from
Tables~\ref{tab:experiment1-main}--\ref{tab:experiment1-modules}.
Direct connections lower ATE relative to auxiliary-submap connections from
$0.505$ to $0.118$\,m on Apartment, $0.175$ to $0.033$\,m on Cubicle,
and $3.370$ to $1.508$\,m on Office. They also improve on Anchor-only in
Apartment and Cubicle, avoiding the auxiliary pipeline's large ATE increases
for these fixed loop pairs.

\noindent\textit{End-to-end loop comparison.}
Fig.~\ref{fig:clio-loop-comparison} compares Baseline and B2 on Cubicle
with Anchor disabled. ATE decreases from 0.2125 to 0.0419\,m (80.3\%).
The pipelines accept four and two loops, respectively, so this is an end-to-end
comparison rather than a fixed-loop ablation. The INS NTU Indoor reconstruction in
Fig.~\ref{fig:indoor-dense-map} further compares Baseline with Full (B3),
showing cleaner inter-segment alignment with the complete system.

\subsection{Runtime Efficiency}

We measure processing throughput on one NVIDIA L20 with dense-map export disabled.
B0 and Full use identical preselected keyframes and the same implementation,
run serially, and
are each repeated three times. On TUM room (186 keyframes), B0 processes
$9.138\pm0.097$ keyframes/s and Full processes $8.026\pm0.095$ keyframes/s;
their processing times are $20.356\pm0.216$ and $23.177\pm0.274$\,s,
respectively. On Clio Apartment (263 keyframes), throughput changes from
$9.240\pm0.093$ to $8.123\pm0.069$ keyframes/s, and processing time from
$28.466\pm0.287$ to $32.377\pm0.276$\,s. Thus the complete anchor and loop
modules add 13.9\% and 13.7\% processing time while retaining more than eight
selected keyframes/s. Peak GPU memory changes from 7468 to 7470\,MiB on TUM
room and from 7472 to $7739\pm230$\,MiB on Clio Apartment. These measurements
are selected-keyframe processing throughput, not raw-video frame rate or
end-to-end online latency; startup and trajectory export are excluded.

\subsection{Scope and Limitations}

The current implementation is keyframe-driven: pose and map updates occur
when a submap is processed. During robot stops or slow motion, image
displacement may remain below the keyframe threshold, leaving an incomplete
submap pending and delaying reconstruction updates. Frames rejected by the
keyframe filter do not receive independent pose estimates. Suppressing
redundant keyframes at rest is expected; the practical limitation is the
absence of continuous per-frame localization between submap updates.
End-to-end online latency has not been benchmarked, and a complete real-time
SLAM system for robotic deployment remains future work.
Outdoor applicability can also be limited when selected views are dominated
by sky or distant, weakly textured regions, which provide insufficient scene
structure for reliable VGGT-based relative-scale inference.

\section{Conclusion}

Fixed-root anchoring lowers no-loop ATE on all seven module-ablation
sequences, while root-restricted SALAD selection improves over the fixed-middle
reference on all four anchor-design sequences. With identical accepted loop
pairs, direct connections outperform auxiliary-submap connections on all three
Clio scenes, avoiding the latter's large ATE increases.

The resulting correction uses one factor per accepted loop between existing
poses; the comparison covers the complete connection pipeline, including its
state representation and robust treatment.

The short-sequence check nevertheless rejects plausible revisits.
The evaluated results support the proposed scale anchoring and direct
connection design. Broader held-out evaluation and independent candidate
annotation remain future work.

\bibliographystyle{IEEEtran}
\bibliography{references}

\begin{thebibliography}{10}
\providecommand{\url}[1]{#1}
\csname url@rmstyle\endcsname
\providecommand{\newblock}{\relax}
\providecommand{\bibinfo}[2]{#2}
\providecommand\BIBentrySTDinterwordspacing{\spaceskip=0pt\relax}
\providecommand\BIBentryALTinterwordstretchfactor{4}
\providecommand\BIBentryALTinterwordspacing{\spaceskip=\fontdimen2\font plus
\BIBentryALTinterwordstretchfactor\fontdimen3\font minus
  \fontdimen4\font\relax}
\providecommand\BIBforeignlanguage[2]{{%
\expandafter\ifx\csname l@#1\endcsname\relax
\typeout{** WARNING: IEEEtran.bst: No hyphenation pattern has been}%
\typeout{** loaded for the language `#1'. Using the pattern for}%
\typeout{** the default language instead.}%
\else
\language=\csname l@#1\endcsname
\fi
#2}}

\bibitem{maggio2026vggtslam2}
D.~Maggio and L.~Carlone, ``{VGGT-SLAM 2.0}: Real time dense feed-forward scene
  reconstruction,'' \emph{arXiv preprint arXiv:2601.19887}, 2026.

\bibitem{wang2025vggt}
J.~Wang, M.~Chen, N.~Karaev, A.~Vedaldi, C.~Rupprecht, and D.~Novotny,
  ``{VGGT}: Visual geometry grounded transformer,'' \emph{arXiv preprint
  arXiv:2503.11651}, 2025.

\bibitem{wang2024dust3r}
S.~Wang, V.~Leroy, Y.~Cabon, B.~Chidlovskii, and J.~Revaud, ``{DUSt3R}:
  Geometric 3d vision made easy,'' \emph{arXiv preprint arXiv:2312.14132},
  2024.

\bibitem{wang2024spann3r}
H.~Wang and L.~Agapito, ``3d reconstruction with spatial memory,'' \emph{arXiv
  preprint arXiv:2408.16061}, 2024.

\bibitem{liu2024slam3r}
Y.~Liu, S.~Dong, S.~Wang, Y.~Yang, Q.~Fan, and B.~Chen, ``{SLAM3R}: Real-time
  dense scene reconstruction from monocular {RGB} videos,'' \emph{arXiv
  preprint arXiv:2412.09401}, 2024.

\bibitem{murai2024mast3rslam}
R.~Murai, E.~Dexheimer, and A.~J. Davison, ``{MASt3R-SLAM}: Real-time dense
  {SLAM} with 3d reconstruction priors,'' \emph{arXiv preprint
  arXiv:2412.12392}, 2024.

\bibitem{maggio2025vggtslam}
D.~Maggio, H.~Lim, and L.~Carlone, ``{VGGT-SLAM}: Dense {RGB} {SLAM} optimized
  on the {$\mathrm{SL}(4)$} manifold,'' \emph{arXiv preprint arXiv:2505.12549},
  2025.

\bibitem{mandal2026vggtslampp}
A.~Mandal, R.~Kumar, S.~S. Harithas, and C.~Arora, ``{VGGT-SLAM++},''
  \emph{arXiv preprint arXiv:2604.06830}, 2026.

\bibitem{deng2025vggtlong}
K.~Deng, Z.~Ti, J.~Xu, J.~Yang, and J.~Xie, ``{VGGT-Long}: Chunk it, loop it,
  align it---pushing {VGGT}'s limits on kilometer-scale long {RGB} sequences,''
  \emph{arXiv preprint arXiv:2507.16443}, 2025.

\bibitem{zhang2026vggtalign}
W.~Zhang, Y.~Wu, S.~Li, and Q.~Wang, ``{VGGT-Align}: Bridging local
  reconstruction and global consistency for long-sequence 3d reconstruction,''
  \emph{arXiv preprint arXiv:2608.15260}, 2026.

\bibitem{strasdat2010scaledrift}
H.~Strasdat, J.~M.~M. Montiel, and A.~J. Davison, ``Scale drift-aware large
  scale monocular {SLAM},'' in \emph{Robotics: Science and Systems}, 2010.

\bibitem{arandjelovic2016netvlad}
R.~Arandjelovi{\'c}, P.~Gronat, A.~Torii, T.~Pajdla, and J.~Sivic, ``{NetVLAD}:
  {CNN} architecture for weakly supervised place recognition,'' in
  \emph{Proceedings of the IEEE Conference on Computer Vision and Pattern
  Recognition}, 2016.

\bibitem{oquab2023dinov2}
M.~Oquab, T.~Darcet, T.~Moutakanni, H.~Vo, M.~Szafraniec, V.~Khalidov,
  P.~Fernandez, D.~Haziza, F.~Massa, A.~El-Nouby, \emph{et~al.}, ``{DINOv2}:
  Learning robust visual features without supervision,'' \emph{arXiv preprint
  arXiv:2304.07193}, 2023.

\bibitem{izquierdo2024salad}
S.~Izquierdo and J.~Civera, ``Optimal transport aggregation for visual place
  recognition,'' in \emph{Proceedings of the IEEE/CVF Conference on Computer
  Vision and Pattern Recognition}, 2024.

\bibitem{campos2021orbslam3}
C.~Campos, R.~Elvira, J.~J.~G. Rodr{\'i}guez, J.~M.~M. Montiel, and J.~D.
  Tard{\'o}s, ``{ORB-SLAM3}: An accurate open-source library for visual,
  visual--inertial, and multimap {SLAM},'' \emph{IEEE Transactions on
  Robotics}, vol.~37, no.~6, pp. 1874--1890, 2021.

\bibitem{milford2012seqslam}
M.~J. Milford and G.~F. Wyeth, ``{SeqSLAM}: Visual route-based navigation for
  sunny summer days and stormy winter nights,'' in \emph{IEEE International
  Conference on Robotics and Automation}, 2012, pp. 1643--1649.

\bibitem{sunderhauf2012switchable}
N.~S{\"u}nderhauf and P.~Protzel, ``Switchable constraints for robust pose
  graph {SLAM},'' in \emph{2012 IEEE/RSJ International Conference on
  Intelligent Robots and Systems}, 2012, pp. 1879--1884.

\bibitem{agarwal2013dcs}
P.~Agarwal, G.~D. Tipaldi, L.~Spinello, C.~Stachniss, and W.~Burgard, ``Robust
  map optimization using dynamic covariance scaling,'' in \emph{IEEE
  International Conference on Robotics and Automation}, 2013.

\bibitem{sturm2012tum}
J.~Sturm, N.~Engelhard, F.~Endres, W.~Burgard, and D.~Cremers, ``A benchmark
  for the evaluation of {RGB-D SLAM} systems,'' in \emph{IEEE/RSJ International
  Conference on Intelligent Robots and Systems}, 2012.

\bibitem{shi2020openloris}
X.~Shi, D.~Li, P.~Zhao, Q.~Tian, Y.~Tian, Q.~Long, C.~Zhu, J.~Song, F.~Qiao,
  L.~Song, Y.~Guo, Z.~Wang, Y.~Zhang, B.~Qin, W.~Yang, F.~Wang, R.~H.~M. Chan,
  and Q.~She, ``Are we ready for service robots? the {OpenLORIS-Scene} datasets
  for lifelong {SLAM},'' in \emph{IEEE International Conference on Robotics and
  Automation}, 2020.

\bibitem{maggio2024clio}
D.~Maggio, Y.~Chang, N.~Hughes, M.~Trang, D.~Griffith, C.~Dougherty,
  E.~Cristofalo, L.~Schmid, and L.~Carlone, ``Clio: Real-time task-driven
  open-set 3d scene graphs,'' \emph{IEEE Robotics and Automation Letters},
  vol.~9, no.~10, pp. 8921--8928, 2024.

\bibitem{sarlin2022lamar}
P.-E. Sarlin, M.~Dusmanu, J.~L. Sch{\"o}nberger, P.~Speciale, L.~Gruber,
  V.~Larsson, O.~Miksik, and M.~Pollefeys, ``{LaMAR}: Benchmarking localization
  and mapping for augmented reality,'' in \emph{European Conference on Computer
  Vision}, 2022.

\bibitem{cortes2018advio}
S.~Cort{\'e}s, A.~Solin, E.~Rahtu, and J.~Kannala, ``{ADVIO}: An authentic
  dataset for visual-inertial odometry,'' in \emph{European Conference on
  Computer Vision}, 2018.

\bibitem{deng2025mneslam}
T.~Deng \emph{et~al.}, ``{MNE-SLAM}: Multi-agent neural {SLAM} for mobile
  robots,'' in \emph{Proceedings of the IEEE/CVF Conference on Computer Vision
  and Pattern Recognition}, 2025, pp. 1485--1494.

\end{thebibliography}

\end{document}